\documentclass[10pt,twocolumn,letterpaper]{article}

\usepackage{btas}
\usepackage{times}
\usepackage{epsfig}
\usepackage{graphicx}
\usepackage{amsmath}
\usepackage{amssymb}
\graphicspath{{images/}}
\usepackage{subfig}
\usepackage{epstopdf}
\usepackage{multirow}
\DeclareMathOperator*{\argmin } {arg\, min}

\usepackage{url}
\usepackage{array}
\newcolumntype{M}[1]{>{\centering\arraybackslash}m{#1}}
\newcolumntype{P}[1]{>{\centering\arraybackslash}p{#1}}
\usepackage{multirow}
\usepackage{enumitem}

\btasfinalcopy % *** Uncomment this line for the final submission

\ifbtasfinal\fi
\begin{document}

%%%%%%%%% TITLE
%\title{Expression Classification in Children using Mean Supervised Deep Boltzmann Machine}
\title{On Matching Skulls to Digital Face Images: A Preliminary Approach}

\author{Shruti Nagpal$^1$, Maneet Singh$^1$, Arushi Jain$^1$, Richa Singh$^{1,2}$, Mayank Vatsa$^{1,2}$, Afzel Noore$^2$\\
$^1$IIIT Delhi, India, $ ^2$West Virginia University\\
\{\tt\small shrutin, maneets, arushi13023, rsingh, mayank\}@iiitd.ac.in, afzel.noore@mail.wvu.edu
}

\maketitle
\thispagestyle{empty}

%%%%%%%%% ABSTRACT

\begin{abstract}
Forensic application of automatically matching skull with face images is an important research area linking biometrics with practical applications in forensics. It is an opportunity for biometrics and face recognition researchers to help the law enforcement and forensic experts in giving an identity to unidentified human skulls. It is an extremely challenging problem which is further exacerbated due to lack of any publicly available database related to this problem. This is the first research in this direction with a two-fold contribution: (i) introducing the first of its kind skull-face image pair database, IdentifyMe, and (ii) presenting a preliminary approach using the proposed semi-supervised formulation of transform learning. The experimental results and comparison with existing algorithms showcase the challenging nature of the problem. We assert that the availability of the database will inspire researchers to build sophisticated skull-to-face matching algorithms.

\end{abstract}

%%%%%%%%% BODY TEXT

\section{Introduction}

\vspace{6pt}

\hrule
\vspace{5pt}
\textit{``Some say, `That's such an old case, why do it?' I say, `Why not?' They still deserve their names back."} - Joe Mullins, National Center for Missing and Exploited Children.
\vspace{5pt}
\hrule
\vspace{10pt}

\begin{figure}[!t]
\centering
\includegraphics[width=7cm]{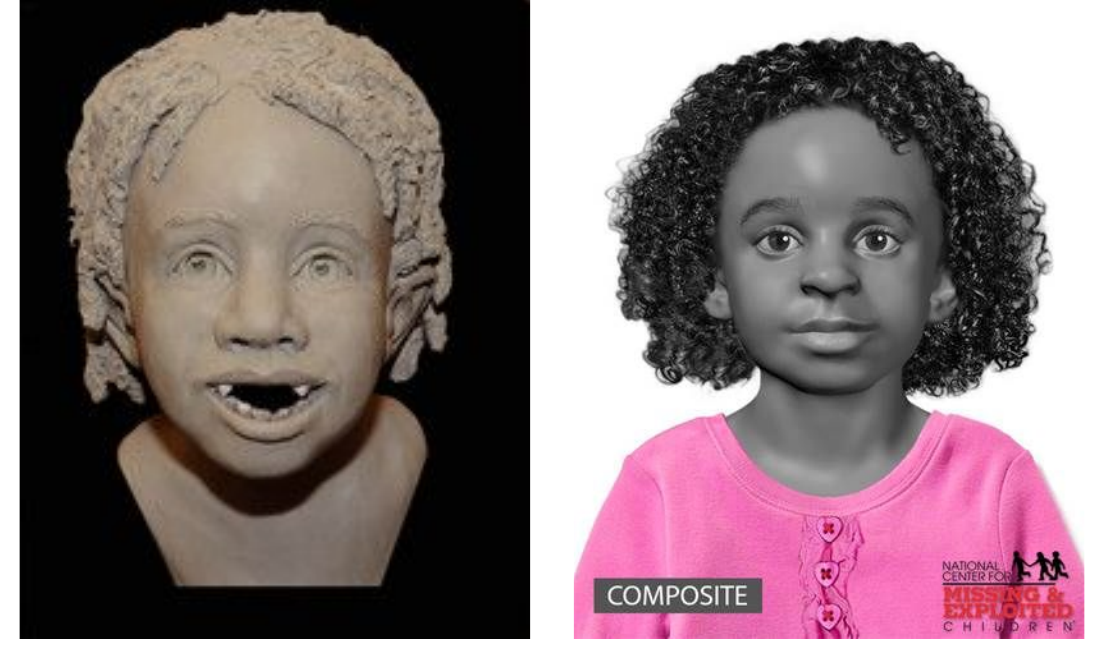}
\caption{Clay reconstruction and 3D facial composite of a girl whose remains were found in Opelika, Oklahoma.}
\label{fig:unsolved}
\end{figure}

In 2012, law enforcement authorities found the remains of a young girl in Opelika, Oklahoma\footnote{https://identifyus.org/en/cases/9834}. From the skull of the girl, forensic experts reconstructed the face (top left in Figure \ref{fig:unsolved}) and released it to the public. After few years, they also created the computerized composite of the girl and shared it with the public. However, more than five years have passed and the identity of the girl is still unknown.

\begin{figure}[!t]
\centering
\includegraphics[width=7cm]{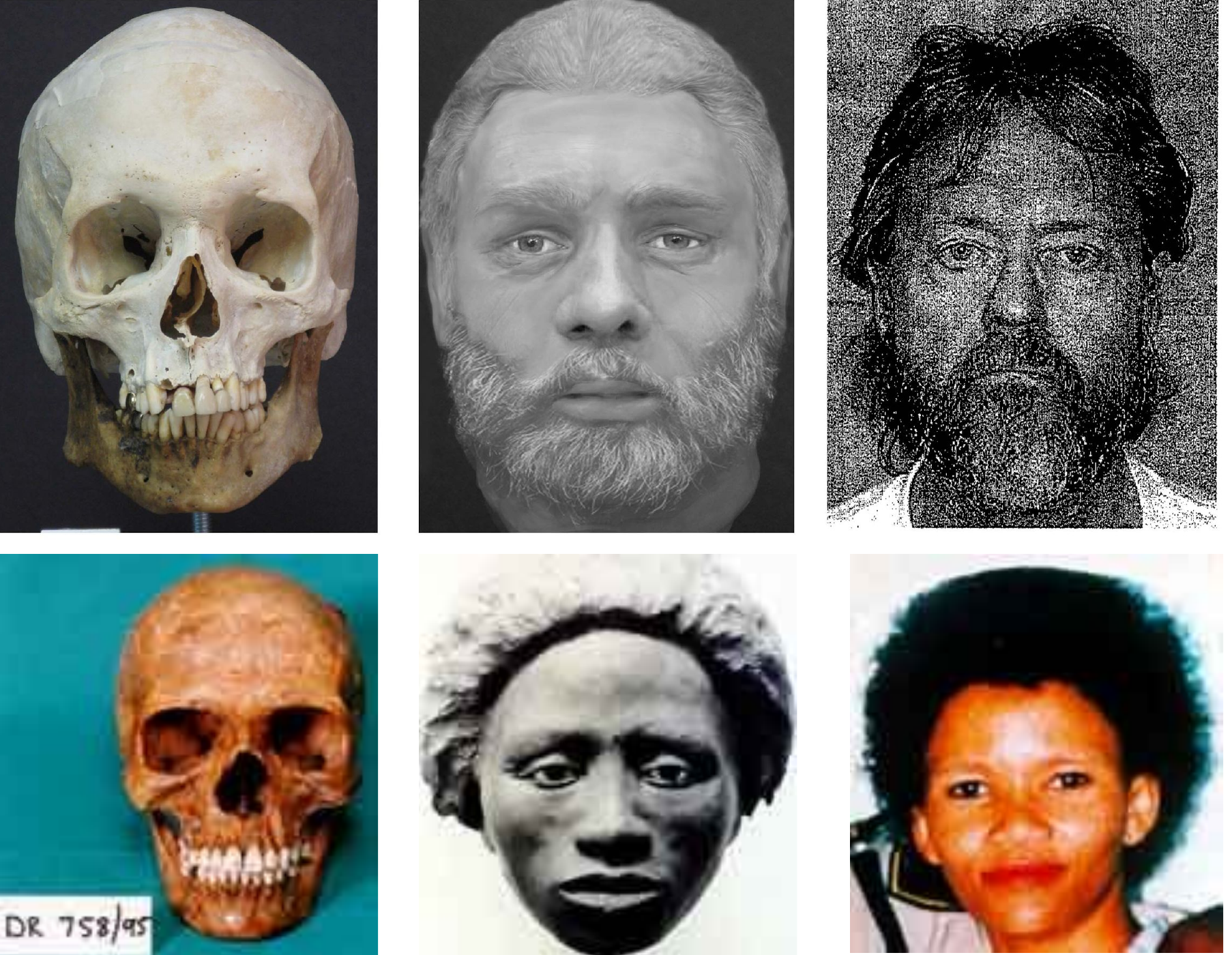}
\caption{Skeleton, reconstructed clay composite, and actual face image of two individuals. The images are taken from the Internet.}
\label{fig:intro}
\end{figure}

This case and several other such cases show that identification is not only required by the living people but it is also important for the dead. Identification for the living has been well studied and multiple biometric modalities provide efficient and accurate methods for identification. However, dead bodies and remains of individuals who have suffered unfortunate death due to crime or natural calamities, often do not have the luxury of having biometrics modalities intact. Fingerprints \cite{fingerprint} and iris \cite{iris} may be decomposed while dental and DNA patterns may not exist in medical or national identification databases. Further, there are several cases where only some body parts are available, for instance head, hand or leg. While the bones in hand or leg can be used to determine DNA and, to a certain extent ethnicity and gender as well, however, forensic experts can reconstruct a lot more information if the skull is available. They can determine the DNA, age, gender (depending on the age), and reconstruct the entire face which may be very useful in forensic and law enforcement investigation. %These instances showcase the effectiveness of Joe Mullin's statement \textit{``All the information about you that makes you the individual that you are is etched into your skull."}. 

This research paper explores how the biometrics and face recognition community can help forensic experts in identifying the person from the skull. Given the skull of a person, forensic experts first determine the age, gender and ethnicity of the person using several morphological measures  \cite{gender}, followed by generating the facial reconstruction from the skull using clay sculpture. This process is generally performed manually; however, recently some of the labs have started using 3D reconstruction software such as FaceIT and ReFace \cite{whoIsThis}. Figure \ref{fig:intro} shows the skeleton, clay reconstruction, and original face images (from left to right) from two solved cases\footnote{http://catyanaskoryfalsetti.com/forensic-art/} \footnote{https://archives.fbi.gov/archives/about-us/lab/forensic-science-communications/fsc/jan2001/phillips.htm}. This process, when performed manually, takes a little over a week for one skull while generating computerized 3D reconstruction takes 3-4 days. Given this process, face matching can be performed in two ways: (i) matching the skull with face images, and (ii) matching the composite with face images. Matching reconstructed composite with face images might appear easier, however it requires an expert forensic anthropologist and at least 3-4 days for generating the composite. 

In this research, we examine the feasibility of matching skull to face images, that may be available in a missing person database. We postulate that matching skull to face images can be considered as another dimension/covariate in heterogeneous face recognition research with forensic applications. The first and foremost requirement of pursuing this research is a database of skull and corresponding face images. Therefore, we first prepared the \textit{IdentifyMe - Skull and Face Image Database} which contains mated pairs of skull and face images from 35 individuals and 429 independent skull images. The database will be released to the research community to promote further research in this area. We next propose a preliminary algorithm using transform learning to match skull with face images, along with establishing the baseline skull-to-face matching performance of several existing algorithms. The results show that the problem at hand is a challenging yet feasible research problem, requiring creation of a larger database and specifically designed algorithms to correctly determine the identity of the person with high accuracy/precision.

\section{Proposed IdentifyMe Dataset}

While existing research has focused on creating facial reconstructions from skulls, no work has been done to automate the matching of skull images to digital face images. This is the first work which aims to match a given skull image to a large number of face images to retrieve the identity of the given skull image. Due to the lack of an existing dataset for the above problem, we have prepared the proposed \textit{IdentifyMe} dataset. The proposed dataset contains 464 skull images, and is divided into two parts: (i) skull and digital face image pairs, and (ii) unlabeled supplementary skull images. The first component consists of pairs collected from real world examples of solved skull identification cases, while the second part contains unlabeled skull images. Details of the two components of the proposed IdentifyMe dataset are provided below:

\subsection{Skull and Face Image Pairs}

\begin{figure}
\centering
\subfloat[The first row corresponds to the digital face images, while the second contains their respective skull images.]{{\includegraphics[width=3.2in]{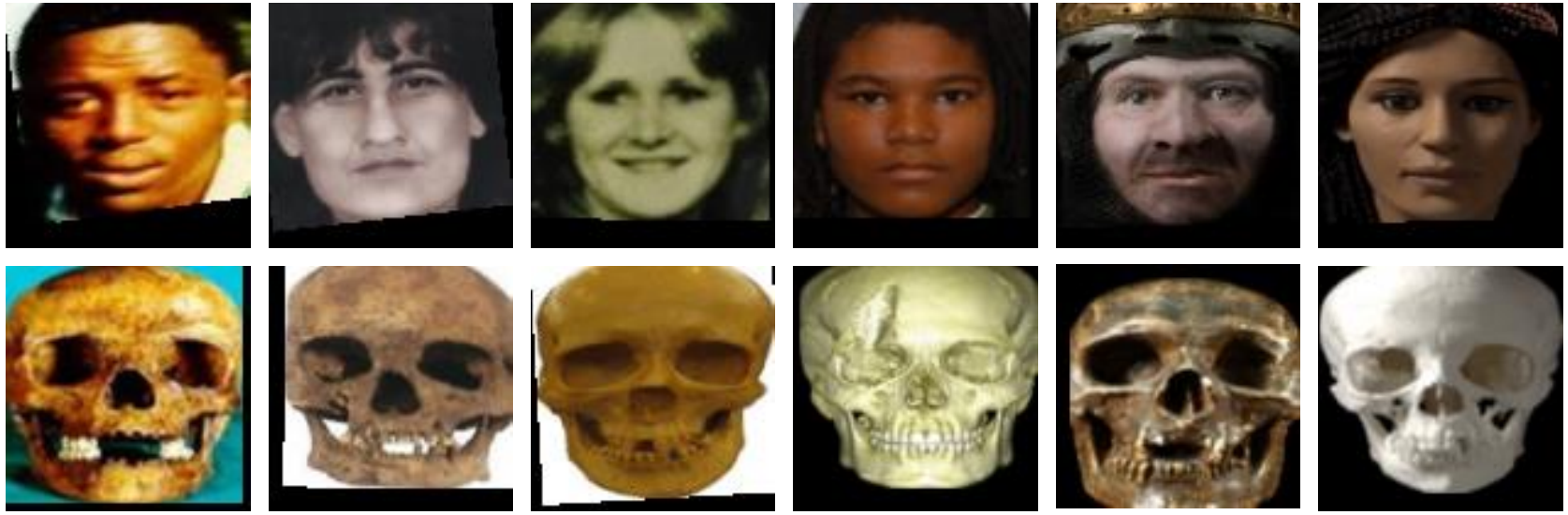} }}

\subfloat[Unlabeled supplementary skull images.]{{\includegraphics[width=3.2in]{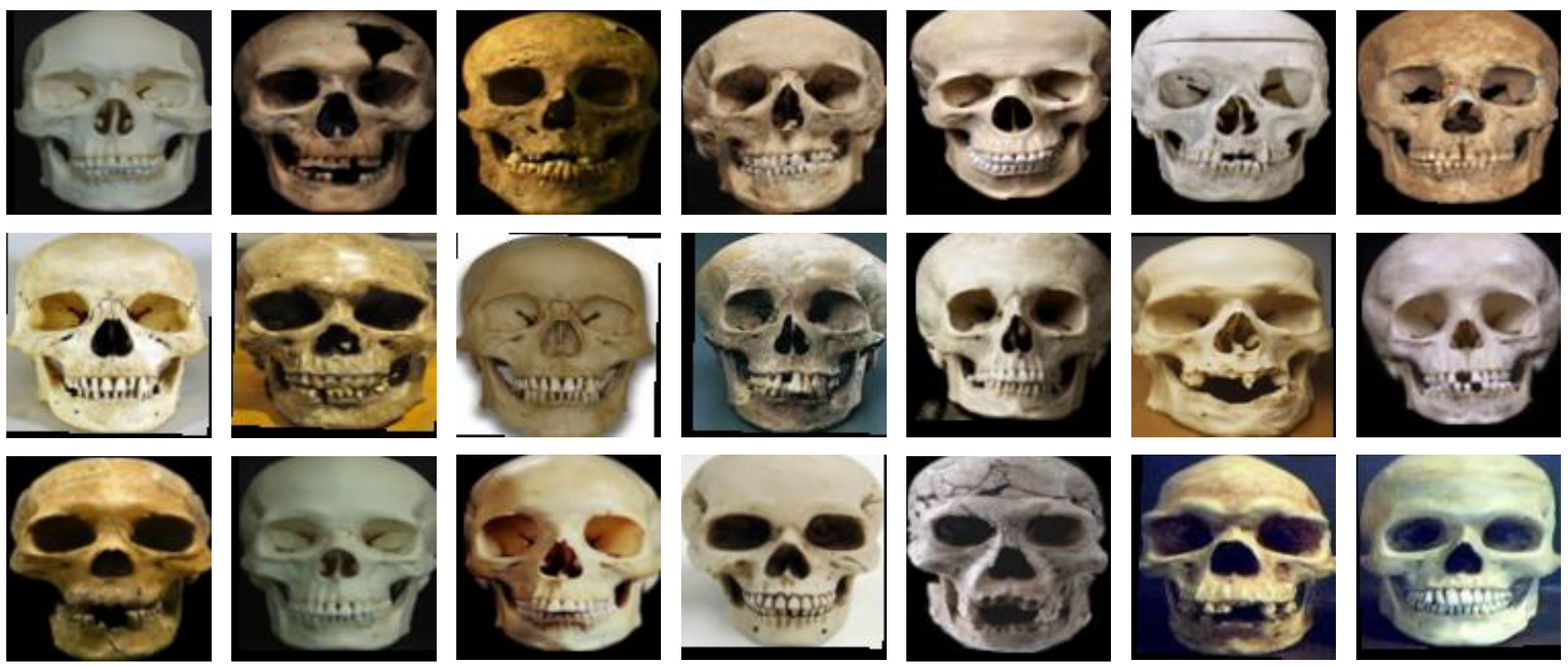} }}%
\caption{Sample images from the proposed IdentifyMe - Skull and Face Image Database. }%
\label{fig:dbase}%
\end{figure}

A total of 35 skull images and their corresponding face images are collected from various sources. Some of these pairs correspond to real world cases where a skull was found and later identified to belong to a missing person. Owing to the unavailability of such kind of real world data, this component also consists of some famous reconstructed face and skull image pairs. Figure \ref{fig:dbase}(a) presents some sample skull-digital face image pairs from the proposed IdentifyMe dataset. Unavailability of high quality, well-illuminated digital face images for real world cases further renders the given problem challenging. 

%\begin{figure}
%\centering
%\includegraphics[width=3.2in]{pairs.pdf}
%\caption{Sample skull and digital face image pairs. The first row corresponds to the digital face images, while the second contain contain their respective skull images. }
%\label{fig:fig1}
%\end{figure}
%

%\begin{figure*}
%\centering
%\includegraphics[width=3.2in]{pairs.pdf}
%\includegraphics[width=6in]{db.pdf}
%\caption{Sample unlabeled supplementary skull images.}
%\label{fig:fig2}
%\end{figure*}

\subsection{Unlabeled Supplementary Skull Images}
The second component of the proposed dataset consists of 429 unlabeled, supplementary skull images. While it is difficult to obtain skull-digital image pairs, however, one can also obtain unlabeled skull images from the Internet. Although the identities of the skull images of this component are unknown, these set of images can be utilized for unsupervised learning based algorithms. Owing to the heterogeneity of the problem, and large variations observed between the domains of digital face images and skulls, this component of unlabeled supplementary skull images can thus be used to reduce the domain gap between the two. 

The proposed IdentifyMe dataset will be made publicly available to the research community\footnote{http://iab-rubric.org/resources/identifyme.html}. Moreover, this dataset will also be made available as a \textit{contributory} dataset, wherein researchers are invited to add to the existing dataset by providing some of the images (obtained from the Internet or other sources). We will first ensure that the images are not already available in the database and then release the new set of images as part of the database with acknowledgment to the contributors. This will ensure incremental growth in the resources and foster research in the domain of skull to digital face image matching.  

%As a part of this research, we also provide a supplementary set of skull images only. This dataset consists of skull images taken from the internet. In all, the dataset comprises of 429 skull images which can be used for the purpose of unsupervised training.  

\subsection{Protocols}
\label{sec:protocol}

In order to standardize further research on skull to digital photo matching, two protocols are provided for the proposed dataset. In real world scenarios of missing person, it is unlikely for law enforcement agencies to perform face verification (1:1 matching) with skull images. Therefore, two identification protocols are defined on the proposed dataset. Each protocol consists of five-fold cross validation in order to remove any bias of a particular train-test partition. As is the case in real world, the digital photos correspond to the gallery images (or the database), and the skull images correspond to the probes. The two protocols are defined as follows:

\noindent \textbf{Protocol-1:} The first protocol utilizes only the 35 skull-face image pairs. These pairs are divided into five folds for performing five fold cross validation. Each fold consists of seven pairs, such that four folds are used for training, and the remaining one fold is used for testing. %Therefore, each fold consists of seven digital images, and seven skull images. 

\noindent \textbf{Protocol-2:} In real world scenarios, a skull image is required to be identified against a large database of face images from different gender, age and ethnicity. Therefore, in order to mimic this real world scenario, the second protocol consists of an extended gallery, where 993 additional digital images are added in the gallery (to obtain a gallery of 1000 images). Therefore, each fold in this protocol consists of 1000 gallery images (digital face images) and 7 probe images (skull images).

The first protocol attempts to perform identification within the proposed dataset only, while the second protocol mimics the real world scenario of having a much larger dataset against which a given input will be matched. The above defined train-test partitions, for all five folds will be provided along with the dataset to the research community.

\begin{figure*}[!t]
\centering
\includegraphics[width=15cm]{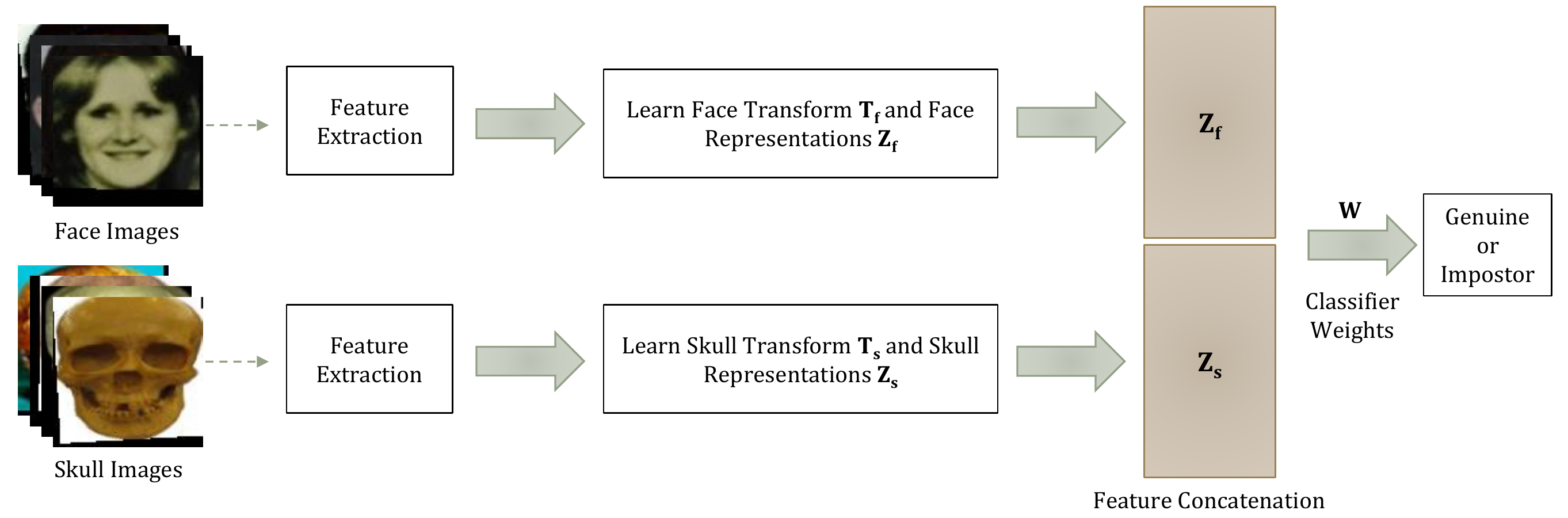}
\caption{Illustrating the steps involved in the proposed Supervised Transform Learning for skull to face image recognition. Input images or features of skulls and face images are used to learn transform matrices ($\mathbf{T_s}, \mathbf{T_f}$) and corresponding representations ($\mathbf{Z_s}, \mathbf{Z_f}$). The learned representations are then used to learn the classifier weights $\mathbf{W}$.}
\label{fig:algo}
\end{figure*}

\section{Proposed Algorithm}

For skull to face matching, there are two major restrictions in building automated algorithms: (i) limited training samples per subject and (ii) overall limited training data. Due to the sensitive nature of the problem, getting multiple training samples from each subject is exceptionally challenging. The problem is further exacerbated with availability of limited labeled training data (specifically, no training data before this research). Therefore, data intensive algorithms such as deep learning will be extremely demanding and may overfit. On the other hand, algorithms which do not require training such as handcrafted features or relatively less training such as dictionary learning or transform learning can be explored. In this research, we propose transform learning based skull to face matching by utilizing the domain specific knowledge (i.e. face) and problem specific knowledge (i.e. skull).

Ravishankar and Bresler \cite{Ravishankar13} proposed transform learning, an analysis equivalent of dictionary learning. It is a representation learning technique to learn a transformation space, $\mathbf{T}$ which is used to produce a representation, $\mathbf{Z}$ for given input samples, $\mathbf{X}$. It is mathematically written as:
 \begin{equation} \label{transformNew}
\begin{gathered}
\argmin_{\textit{$\mathbf{T, Z}$}} \left \|\mathbf{TX} - \mathbf{Z} \right \|_{F}^{2} +  \lambda\  \big( \epsilon \left \|\mathbf{T} \right \|_{F}^{2}  - log \det \mathbf{T} \big) \\
s.t. \left \|\mathbf{Z}\right \|_{0} \leq \tau
\end{gathered}
\end{equation}
A penalty term, in the form of $\ell_{2}$-norm regularization of the transformation matrix, $\mathbf{T}$ is added  in order to balance the scale. The ($log \det \mathbf{T}$) term is the log-determinant regularizer \cite{logdet}, added to prevent possible degenerate solutions. Existing papers  \cite{RavishankarSIAM, Ravishankar15_2} show the following alternating minimization approach to learn $\mathbf{T}$ and $\mathbf{Z}$:

\begin{enumerate}
\item Initialize $\mathbf{T}$, $\mathbf{Z}$
\item Fix $\mathbf{T}$, learn $\mathbf{Z}$ as:
\begin{equation} \label{solve}
{
\mathbf{Z} \leftarrow \argmin_{\textit{$\mathbf{Z}$}} \left \|\mathbf{TX} - \mathbf{Z} \right \|_{F}^{2},\ such\ that \  \left \|\mathbf{Z}\right \|_{0} \leq \tau
}
\end{equation}
\item Fix $\mathbf{Z}$, learn $\mathbf{T}$ as:
\begin{equation} \label{solveT}
{\mathbf{T} \leftarrow \argmin_{\textit{$\mathbf{T}$}} \left \|\mathbf{TX} - \mathbf{Z} \right \|_{F}^{2} +  \lambda\  \big( \epsilon \left \|\mathbf{T} \right \|_{F}^{2}  - log \det \mathbf{T} \big) 
}
\end{equation}
\end{enumerate}

Natively, transform learning is an unsupervised representation learning approach. In this research, we first extend the formulation and propose Semi-Supervised Transform Learning. We next propose two algorithms for skull to face identification: (a) using Unsupervised Transform Learning (US-TL) and (b) using Semi-Supervised Transform Learning (SS-TL).

%In \cite{Ravishankar13}, an analysis equivalent of dictionary learning, termed as transformation learning is proposed. It analyzes the data by learning a transformation or basis to produce coefficients. Mathematically, for input data $\mathbf{X}$, it can be expressed as:
%\begin{equation} \label{transform}
%{
%\argmin_{\textit{$\mathbf{T, Z}$}} \left \|\mathbf{TX} - \mathbf{Z} \right \|_{F}^{2},\ such\ that \left \|\mathbf{Z}\right \|_{0} \leq \tau
%}
%\end{equation}
%where, $\mathbf{T}$ and $\mathbf{Z}$ are the transform and the coefficients, respectively. Relating transformation learning to the dictionary learning formulation in Equation \ref{KSVD}, it can be seen that dictionary learning is an inverse problem while transformation learning is a forward problem. In order to avoid the degenerate solutions of Equation \ref{transform}, Ravishankar and Bresler \cite{Ravishankar13} proposed the following formulation: 
%
%\begin{equation} \label{transformNew}
%\begin{gathered}
%\argmin_{\textit{$\mathbf{T, Z}$}} \left \|\mathbf{TX} - \mathbf{Z} \right \|_{F}^{2} +  \lambda\  \big( \epsilon \left \|\mathbf{T} \right \|_{F}^{2}  - log \det \mathbf{T} \big) 
%s.t. \left \|\mathbf{Z}\right \|_{0} \leq \tau
%\end{gathered}
%\end{equation}

\subsection{Skull to Face Matching using Unsupervised Transform Learning (US-TL)}
The proposed unsupervised transform learning based algorithm utilizes the traditional transform learning model to learn representations for the face images and skull images independently. The proposed algorithm is explained as follows:

\begin{enumerate}

\item Given input training face images $\mathbf{X_f}$, a transformation matrix, $\mathbf{T_f}$ and representation $\mathbf{Z_f}$ are learned.
\begin{equation} \label{face}
\argmin_{\textit{$\mathbf{T_f, Z_f}$}} \left \|\mathbf{T_{f}X_{f}} - \mathbf{Z_{f}} \right \|_{F}^{2} +\ \lambda ( \epsilon \left \|\mathbf{T_f}\right \|_{F}^{2} -\ log \det \mathbf{T_f})
\end{equation}
Equation \ref{face} is learned using unlabeled face images with the aim of learning effective representations in a transformed space. 
\item Similarly, a transform learning model is trained for skull images $\mathbf{X_s}$ to learn representations $\mathbf{Z_s}$ and transformation matrix, $\mathbf{T_s}$:
\begin{equation} \label{skull}
\argmin_{\textit{$\mathbf{T_s, Z_s}$}} \left \|\mathbf{T_{s}X_{s}} - \mathbf{Z_{s}} \right \|_{F}^{2} +\ \lambda ( \epsilon \left \|\mathbf{T_s}\right \|_{F}^{2} -\ log \det \mathbf{T_s})
\end{equation}
In order to learn $\mathbf{T_s}$ and $\mathbf{Z_s}$, unlabeled skull images are used (details are given in implementation details). 
%However, due to the limited training data of the skull images, the transformation matrix $\mathbf{T_s}$ of the model is pre-trained with unlabeled face images. 
\item Euclidean distances of the final representations, $\mathbf{Z_f}$ and $\mathbf{Z_s}$ are used to perform identification or match skull images to face images.
\end{enumerate}

\subsection{Skull to Face Matching using Semi-Supervised Transform Learning (SS-TL)}
The unsupervised transform learning approach does not specifically encode ``matchability'' and ``supervision'' with respect to domain specific knowledge (i.e. face) and problem specific knowledge (i.e. skull). We extend the unsupervised transform learning formulation to semi-supervised transform learning for matching a given skull image to faces. Figure \ref{fig:algo} illustrates the overview of the proposed algorithm. 

In this research, the unsupervised transform learning is extended using a perceptron like classifier-weight learning. For face and skull input data ($\mathbf{X_f}$ and $\mathbf{X_s}$), the modified formulation is expressed as: 
\begin{equation} \label{final}
\begin{gathered}
\argmin_{\textit{$\mathbf{T_s, Z_s, T_f, Z_f, W}$}}  \left \|\mathbf{T_{f}X_{f}} - \mathbf{Z_{f}} \right \|_{F}^{2} \\ 
+ \ \lambda ( \epsilon \left \|\mathbf{T_f}\right \|_{F}^{2} -\ log \det \mathbf{T_f}) + \left \|\mathbf{T_{s}X_{s}} - \mathbf{Z_{s}} \right \|_{F}^{2} \\ 
+ \ \lambda ( \epsilon \left \|\mathbf{T_s}\right \|_{F}^{2} -\ log \det \mathbf{T_s}) + \mathbf{W} [\mathbf{Z_{f}Z_{s}}] 
\end{gathered}
\end{equation}

\noindent where, $\mathbf{W} [\mathbf{Z_{f}Z_{s}}]$ is the supervision term which enables supervision by encoding class information to perform improved matching of skull to face representations, and $\left \|\mathbf{T_{f}X_{f}} - \mathbf{Z_{f}} \right \|_{F}^{2}$ and $\left \|\mathbf{T_{s}X_{s}} - \mathbf{Z_{s}} \right \|_{F}^{2}$ terms are used to learn representations. Since supervised training samples are very limited, we further extend the training process which utilizes both unsupervised and supervised samples as follows: 

\begin{enumerate}
\item The first step in the proposed algorithm involves learning two independent representations  and transforms ($\mathbf{Z_f}$, $\mathbf{Z_s}$, $\mathbf{T_f}$, and $\mathbf{T_s}$) using Equations \ref{face} and \ref{skull} for face and skull, respectively. This is performed using unlabeled face and skull data. 

\item  In Equation \ref{final}, initialize $\mathbf{T_f, Z_f}$ and $\mathbf{T_s, Z_s}$ using the values obtained from unlabeled training data obtained in Step 1. Initialize $\mathbf{W}$ with ones. 

\item Using labeled training data, recompute $\mathbf{T_s, Z_s, T_f, Z_f}$ and compute $\mathbf{W}$. 

\end{enumerate}

%In the next step, t The representations are then concatenated and a \textit{linear perceptron weight matrix}, $\mathbf{W}$ is learned, which incorporates supervision in order to perform matching. This step is done for It is mathematically expressed as follows:

%\begin{equation} \label{Face}
%\argmin_{\textit{$\mathbf{T_f}$}} \left \|\mathbf{T_{f}X_{f}} - \mathbf{Z_{f}} \right \|_{F}^{2} +\ \lambda ( \epsilon \left \|\mathbf{T_f}\right \|_{F}^{2} -\ log \det \mathbf{T_f})
%\end{equation}
%
%\begin{equation} \label{skull}
%\argmin_{\textit{$\mathbf{T_s}$}} \left \|\mathbf{T_{s}X_{s}} - \mathbf{Z_{s}} \right \|_{F}^{s} +\ \lambda ( \epsilon \left \|\mathbf{T_s}\right \|_{F}^{2} -\ log \det \mathbf{T_s})
%\end{equation}

In the proposed semi-supervised training, we utilize both domain specific and problem specific knowledge and later incorporate supervision for improved performance. To train the model, alternating minimization approach is utilized.

\subsection{Implementation Details}

\noindent \textbf{Unsupervised Transform Learning:} Face images from CMU Multi-PIE \cite{mpie} dataset have been used as unlabeled face images to train Equation \ref{face} and learn $\mathbf{T_f}$ and $\mathbf{Z_f}$. The same images are also used for pre-training the model in Equation \ref{skull}. Unlabeled Supplementary Skull images from the proposed \textit{IdentifyMe} dataset are used to perform unsupervised training to learn $\mathbf{T_s}$ and $\mathbf{Z_s}$.
%\vspace{3pt}

\noindent \textbf{Semi-Supervised Transform Learning:}  $\mathbf{T_f}$, $\mathbf{Z_f}$ and  $\mathbf{T_s}$, $\mathbf{Z_s}$ are learned using Equation \ref{face} and \ref{skull} as explained above in an unsupervised manner for face and skull images respectively. To incorporate supervision and learn the weight matrix, $\mathbf{W}$, the training partition of the skull and face image pairs from the proposed \textit{IdentifyMe} dataset is utilized.
%\vspace{3pt}

\noindent \textbf{Detection and Registration}: Digital faces are detected using Haar Cascade approach \cite{haar} whereas skulls are manually detected. Area-based alignment is performed using the Image Alignment Toolbox\footnote{http://iatool.net/}, in order to register the skull and face images. In the proposed algorithms, we have used pixel values for $\mathbf{X_f}$ and $\mathbf{X_s}$ (represented as US-TL+Pixels) and SS-TL + Pixels), and HOG features (represented as US-TL + HOG and SS-TL + HOG). Experiments are performed on a workstation with Xeon 2.7GHz processor and 64GB RAM under MATLAB environment. 

\begin{figure*}
	\captionsetup[subfigure]{labelformat=empty}

    \centering
    \subfloat[]{\includegraphics[clip, trim=0cm 0.6cm 0cm 0cm, width=3.4in]{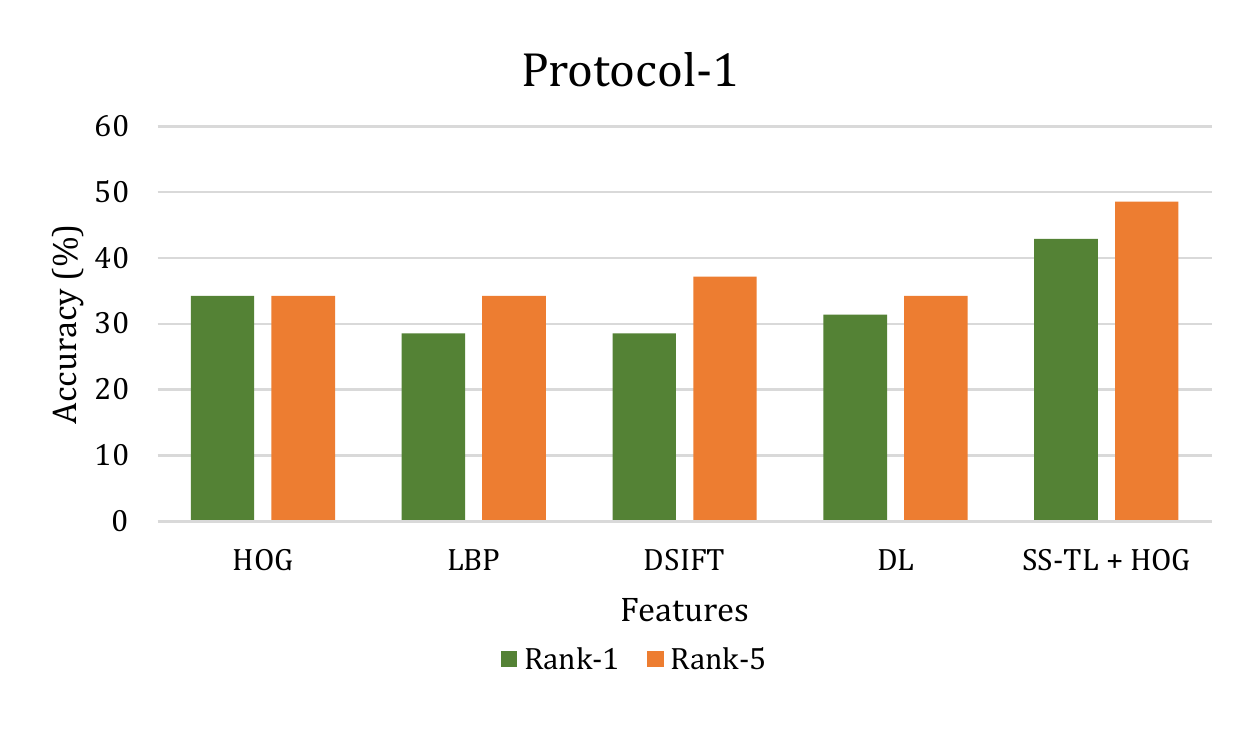} }
    \hspace{-10pt}
    \subfloat[]{\includegraphics[clip, trim=0cm 0.6cm 0cm 0cm, width=3.4in]{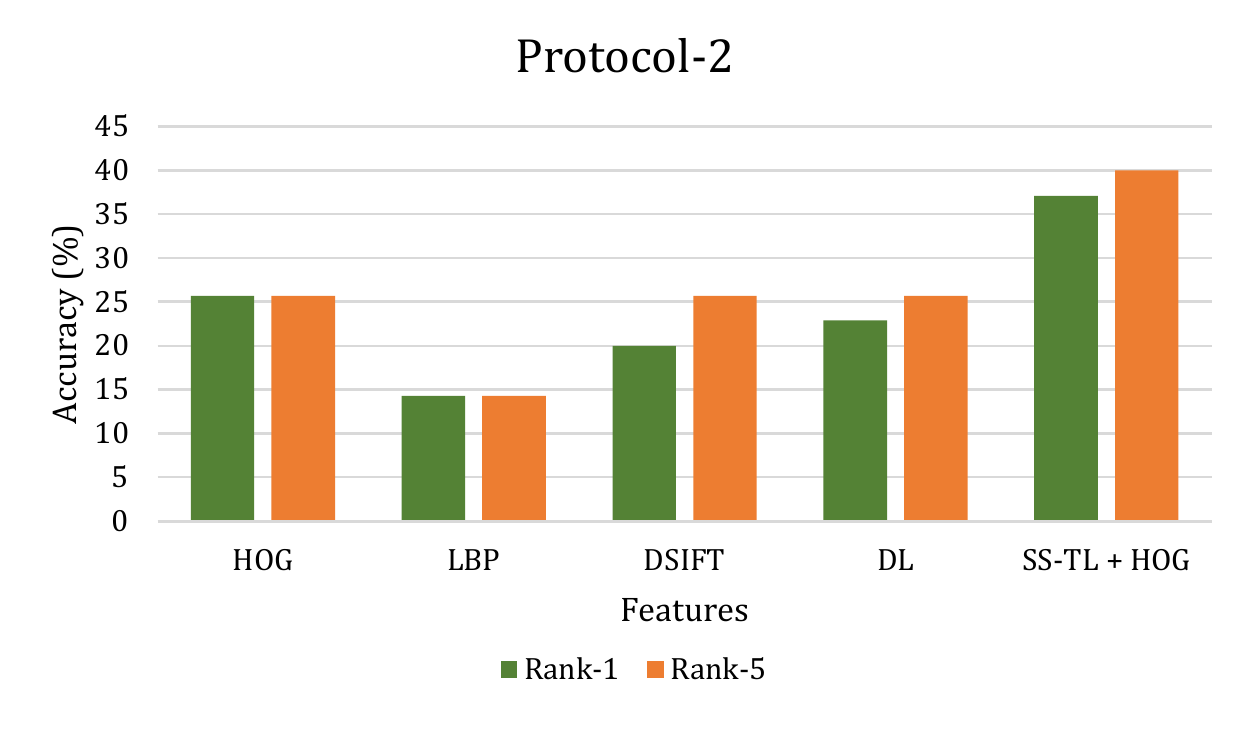} }
    \vspace{-10pt}
    \caption{Bar graphs displaying the identification accuracies (\%) obtained at rank-1 and rank-5 for the two protocols. }
    \label{fig:cmc}
\end{figure*}

\section{Experiments and Results}
Experiments are performed on the above protocols defined in Section \ref{sec:protocol} to present baseline results on the proposed dataset, along with the performance of the proposed algorithm. Identification results are shown with the following hand-crafted features and learned representations: (i) Histogram of Oriented Gradients (HOG) \cite{hog}, (ii) Local Binary Pattern (LBP) \cite{lbp}, (iii) Dense Scale Invariant Feature Transform (DSIFT) \cite{dsift}, and (iv) Dictionary Learning \cite{dl}. In Dictionary Learning, the dictionary is trained on a subset of CMU Multi-PIE \cite{mpie} face dataset, containing frontal well-illuminated face images. %Since no training is performed for the baseline experiments, the training folds and supplementary unlabeled data are not used. 

Once the features are extracted using the above algorithms, Euclidean distance is utilized for performing identification. For the proposed transform learning algorithm, results are presented with both Unsupervised Transform Learning (US-TL) and Semi-Supervised Transform Learning (SS-TL). Since the proposed models can extract features from any input data, results are shown with raw pixels (images as it is) and HOG features as input. Data augmentation is performed on the gallery images by flipping across the y-axis and varying the brightness. 

\begin{figure}[!t]
\centering
\includegraphics[clip, trim=2cm 7cm 2cm 7.3cm, width=3in]{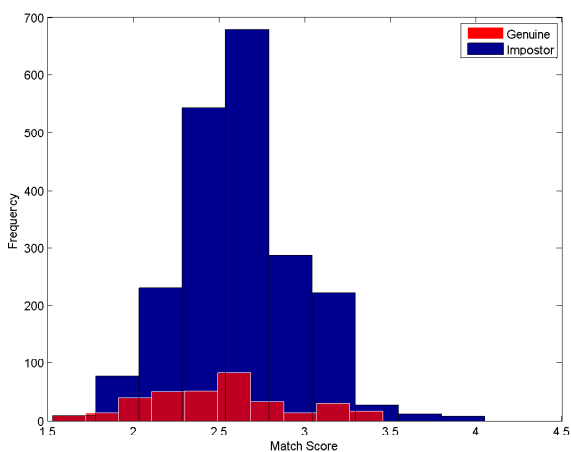} 
\caption{Score distribution of the Euclidean distance values obtained for the genuine and impostor face-skull pairs using HOG features for the first protocol. }
\label{fig:score}
\vspace{-15pt}
\end{figure}

%\begin{table}
%\centering
%\caption{Rank-1 identification accuracies (\%) obtained for Protocol-1 and Protocol-2 on the proposed IdentifyMe Skulls dataset. }
%\begin{tabular}{ |c|c|c|c|c| } 
% \hline
%&  \multicolumn{4}{c|}{\textbf{Features}} \\
% \hline
%  & HOG & LBP & DSIFT & DL \\
% \hline
%\textbf{Protocol-1}  & \textbf{34.2} & 28.5 & 28.5 & 31.4 \\
%\hline
%\textbf{Protocol-2} & \textbf{25.7} & 14.2 & 20.0 & 22.9 \\
%\hline
%\end{tabular}
%\label{tab:rank1}
%\end{table}

%\begin{table}
%\centering
%\caption{Rank-1 identification accuracies (\%) obtained for Protocol-1 and Protocol-2 on the proposed IdentifyMe Skulls dataset using proposed transform learning algorithm. }
%\begin{tabular}{ |c|c|c|c|c| } 
% \hline
%%&  \multicolumn{4}{c|}{\textbf{Features}} \\
% %\hline
% &  \multicolumn{2}{c|}{\textbf{Raw Pixels}} &   \multicolumn{2}{c|}{\textbf{HOG}} \\
% \hline
%  & U-TL& S-TL  & U-TL & S-TL  \\
% \hline
%\textbf{Protocol-1}  & 31.4 & 34.2 & 37.1 & \textbf{42.9} \\
%\hline
%\textbf{Protocol-2} & 25.7 & 28.6 & 34.2 & \textbf{37.1} \\
%\hline
%\end{tabular}
%\label{tab:rank1}
%\end{table}

\begin{table} [t]
\centering
\caption{Rank-1 identification accuracies (\%) obtained for the two protocols on the proposed IdentifyMe dataset.}
\begin{tabular}{ |l|c|c| } 
 \hline
%&  \multicolumn{4}{c|}{\textbf{Features}} \\
 %\hline
 \textbf{Algorithm} & \textbf{Protocol-1} & \textbf{Protocol-2} \\
 \hline
 \hline
 HOG \cite{hog} & 34.2 & 25.7 \\
 \hline
 LBP \cite{lbp} & 28.5 & 14.2 \\
 \hline
DSIFT \cite{dsift} & 28.5 & 20.0 \\
\hline
DL \cite{dl} & 31.4 & 22.9 \\
\hline
\multicolumn{3}{|c|}{\textbf{Proposed Transform Learning}} \\
\hline
US-TL + Pixels & 31.4 & 25.7 \\
\hline
SS-TL + Pixels & 34.2 & 28.6 \\
\hline
US-TL + HOG & 37.1 & 34.2 \\
\hline
\textbf{SS-TL + HOG} & \textbf{42.9} & \textbf{37.1} \\
\hline
\end{tabular}
\label{tab:rank1}
\vspace{-15pt}
\end{table}

Figure \ref{fig:cmc} presents bar graphs summarizing the rank-1 and rank-5 identification accuracies and Table \ref{tab:rank1} tabulates the rank-1 identification accuracies for the two protocols on the proposed dataset. Average accuracies across the five folds are reported. It is observed that:
\begin{itemize}[leftmargin=*]
\item For the first protocol, where the gallery consists of only seven face images corresponding to the seven skull probes, the maximum rank-1 identification accuracy obtained, without using the proposed algorithm is 34.2\% (HOG). Figure \ref{fig:cmc}(a) presents the corresponding bar graph of accuracies with all the features. 

\item The proposed Semi-Supervised Transform Learning (SS-TL) with HOG features as input presents the best performance by achieving a rank-1 identification accuracy of 42.9\% (Table \ref{tab:rank1}). The improvement of more than 8\% over the accuracy achieved by HOG features can directly be attributed to the proposed formulation. 

%Using raw pixels for identification results in a rank-1 identification accuracy of 20\%, which means less than two samples were correctly identified at the first rank. The improvement in accuracies observed upon using HOG over raw pixels motivates the use of image-based features for performing skull to digital image matching. 

\item Table \ref{tab:rank1} also presents the performance of Unsupervised Transform Learning (US-TL) with pixel values and HOG features as input. For both cases, SS-TL performs at least 3\% better than US-TL. This motivates the inclusion of supervision in the proposed model for performing skull to digital image matching. 

\item Figure \ref{fig:score} presents the score distribution of the genuine and impostor pairs using HOG features for the first protocol, obtained across all five folds. No clear distinction can be observed between the scores of the two classes, further motivating the challenging nature of the problem at hand. While the number of genuine scores having large value and impostor scores having low value of Euclidean distance is less, however, the distribution in the middle range is overlapping. 

%\begin{figure} [t]
%\centering
%\includegraphics[width=3.2in]{tp.pdf} 
%\caption{Sample skull-digital image pairs which are correctly identified by almost all algorithms. The first row contains the skull images and the second row contains their corresponding digital face images. }
%\label{fig:tp}
%\end{figure}

\item Figure \ref{fig:cmc}(b) presents the bar graph obtained for the second protocol, where an extended gallery of 993 subjects is also used (i.e, total gallery size is 1000). Similar to the previous protocol, Histogram of Oriented Gradients (HOG) obtains the highest identification accuracy of 25.7\%. Using HOG features as input, the proposed SS-TL model achieves a rank-1 identification accuracy of 37.1\%. The proposed model displays an improvement of more than 10\%, as compared to HOG features based classification.

\item The improved performance of the transform learning models obtained at rank-1 and rank-5 (Figure \ref{fig:cmc}) further motivates learning based algorithms utilizing the supplementary unlabeled component of the proposed IdentifyMe dataset. Since the information content in skulls and digital face images varies significantly, learning from the supplementary unlabeled skull images helps in reducing these variations. 

\item The drop in the best performing rank-1 accuracy from 42.9\% (protocol-1) to 37.1\% (protocol-2) motivates the challenging nature of the problem. In real world cases, where the law enforcement agencies have a large dataset of images against which a given probe image needs to be matched, improved algorithms are required to address this challenging problem.
\end{itemize}

%\begin{figure} [t]
%\centering
%\includegraphics[width=3.2in]{p1Only.pdf} 
%\caption{Sample skull-digital image pairs which are correctly identified in protocol-1, but not in protocol-2.}
%\label{fig:p1}
%\end{figure}

%\item Figure \ref{fig:tp} presents some sample skull-digital image pairs correctly identified at rank-1 by almost all algorithms, for both the protocols. These images indicate that even under the presence of a large gallery (protocol-2, 1000 subjects), face images containing less facial hair can be identified with relative ease. On the other hand, figure \ref{fig:p1} presents some sample images which are correctly identified in protocol-1, and not in protocol-2 by the hand-crafted features. These samples depict the challenging nature of the problem, wherein, under a small sample set, features such as HOG and LBP are able to perform well, however, with the increase in gallery size, there is a requirement of problem-specific algorithms in order to address the same. 

%\item It is also interesting to note that for the second protocol, at the fifth rank, identification using raw pixel information provides better results as compared to other features. This suggests that raw pixels contain some discriminative information, which does not get captured via textural features such as HOG and LBP. This motivates the use of some learning based algorithms, which are able to learn from the pixel values to learn discriminative features, specific to this problem. 

\section{Conclusion}

Determining the identity of human remains, particularly skull, is a long standing problem in forensic applications. Forensic experts first create a facial reconstruction from the skull using clay technique or computerized 3D methods. The composite can then be shared with the media and matched across already available face databases including the missing person database. This research attempts to build an algorithm for matching skull to digital face images. A novel database, \textit{IdentifyMe} database, has been prepared by the authors which consists of more than 450 skull images, including 35 real world skull-face image pairs. Two protocols emulating the real world scenarios, along with their baseline results have also been reported to facilitate research in this direction. Due to the difference in information content of skull and digital images, a Semi-Supervised Transform Learning (SS-TL) algorithm has been proposed for performing identification of skull images. For both the protocols, the proposed SS-TL algorithm achieves improved performance, in comparison with other baseline results. In order to encourage further research in this area, the IdentifyMe dataset will be made publicly available to the research community as a \textit{contributory} dataset, wherein, other researchers would be encouraged to contribute and use the dataset.

\section{Acknowledgment}
S. Nagpal is partially supported through TCS PhD fellowship. M. Vatsa and R. Singh are partially supported through Infosys Center for Artificial Intelligence.

{\small
\bibliographystyle{ieee}
\bibliography{submission_example.bib}
}

\end{document}